\documentclass[letterpaper, 10 pt, conference]{ieeeconf}  

\IEEEoverridecommandlockouts                              

\usepackage{amsmath} 
\usepackage{amssymb}  
\usepackage{dsfont}
\usepackage{graphicx}
\usepackage{subfigure}
\usepackage{psfrag,graphicx,epsfig}
\usepackage{epstopdf}
\usepackage{subfig}
\usepackage{float}
\usepackage{multirow}
\usepackage{pgf,tikz}
\usepackage{xcolor}
\usepackage{siunitx}
\usepackage{color}
\usepackage{flushend}

\title{\LARGE \bf
Generalizable Human-Robot Collaborative Assembly Using \\ Imitation Learning and Force Control
}

\author{Devesh K. Jha, Siddarth Jain, Diego Romeres, William Yerazunis and Daniel Nikovski$^\dag$
\thanks{$^{\dagger}$All authors are with Mitsubishi Electric Research Laboratories (MERL), Cambridge, MA, USA 02139 {\tt\small \{jha,sjain,romeres,yerazunis,nikovski\}@merl.com}}
}

\begin{document}

\maketitle
\thispagestyle{empty}
\pagestyle{empty}

\begin{abstract}

Robots have been steadily increasing their presence in our daily lives, where they can work along with humans to provide assistance in various tasks on industry floors, in offices, and in homes. Automated assembly is one of the key applications of robots, and the next generation assembly systems could become much more efficient by creating collaborative human-robot systems. However, although collaborative robots have been around for decades, their application in truly collaborative systems has been limited. 
This is because a truly collaborative human-robot system needs to adjust its operation with respect to the uncertainty and imprecision in human actions, ensure safety during interaction, etc.
In this paper, we present a system for human-robot collaborative assembly using learning from demonstration and pose estimation, so that the robot can adapt to the uncertainty caused by the operation of humans. Learning from demonstration is used to generate motion trajectories for the robot based on the pose estimate of different goal locations from a deep learning-based vision system. The proposed system is demonstrated using a physical 6 DoF manipulator in a collaborative human-robot assembly scenario. We show successful generalization of the system's operation to changes in the initial and final goal locations through various experiments. 

\end{abstract}

\section{Introduction}\label{sec:introduction}

\begin{figure*}[ht]
    \centering
    \includegraphics[width=1.0\textwidth]{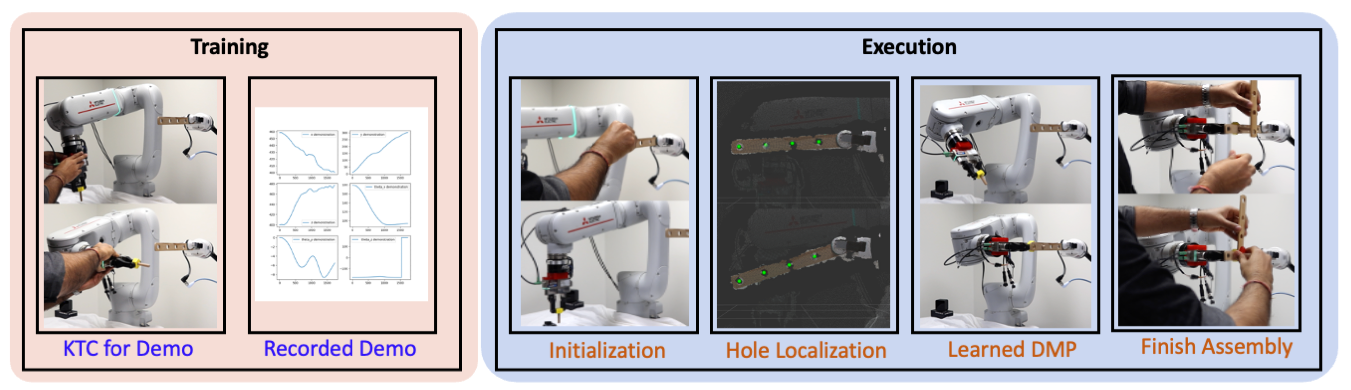} %
    \caption{The overall method presented in this paper for performing collaborative assembly. During training a human provides a direct teaching to the robot to perform insertion using the proposed kinesthetic controller which makes use of the F/T sensor at the wrist of the robot. During execution, the assembly environment is initialized where a bar is grasped by the stationary gripper and a threaded bolt is grasped by the manipulator arm. The robot then performs insertion by using the learned DMP from the demonstration and thus makes the whole system easy to build (as there is no programming involved). This reduces the effort on programming robots using complex motion planning and collision avoidance methods.}
    \label{fig:assembly_steps}
\end{figure*}

The advent and increased adoption of collaborative robots has created possibilities for humans and robots to work together in many applications where barriers previously separated them for safety. One such application of high practical significance is robotic assembly, where executing tasks collaboratively and concurrently might allow for a much more optimal division of labor, such that the robot does more straightforward and repetitive tasks that might also require more physical power, while humans perform more dexterous and intelligent operations. While appealing, such collaborative robotic assembly applications generally cannot be implemented with traditional robot control technology. In traditional robotic assembly (and automation in general), various jigs and fixtures are used to ensure that the assembled parts are exactly where the robot expects to find them so that pre-programmed paths can perform the assembly operation. However, in collaborative assembly, the position of parts will typically vary, depending on where human workers place or hold them. 

Thus, novel technologies for adaptive robot control are needed, where the path of robots is adjusted based on the robotic perception of rich sensory input, typically visual or tactile, similar to how humans work together on a collaborative task while coordinating their actions. In addition, the possibility of safe interaction with collaborative robots allows direct kinesthetic teaching of the operation that is to be performed by the robot by directly holding the manipulator by its end effector and demonstrating the desired trajectory. Suppose one (or few) such demonstrated trajectories can further be adjusted based on sensor input, depending on the variable position of assembled parts and experienced contact forces. In that case, the possibility arises for a fully interactive, programming-less deployment of collaborative robots for collaborative assembly that could be accessible by regular workers in factories without any robotics expertise, thus increasing the usefulness of robotics technology by orders of magnitude. For a flexible human-robot collaborative system, the robot should be able to adjust its motion based on the human's imprecise actions, and this adjustment should be achieved without the laborious use of a motion planning algorithm. Part of meeting these challenges also necessarily involves some understanding and recognition of human intent~\cite{jain2019probabilistic, losey2018review}. Knowing how to recognize and classify human intentions also informs the design of robot behavior, informs aspects of the robot's actions and plans, and contributes to robot interaction that is legible, natural, safe, and comfortable for effective collaboration. While this paper does not attempt to study implicit recognition of human intent, we facilitate collaboration with explicit communication of intentions for collaborative assembly.


This paper presents a method for implementing collaborative human-robot assembly by combining robotic perception and imitation learning. In particular, we present a system where a human and robot collaboratively perform subtasks to assemble parts. In particular, we use a vision system to guide the robotic system to estimate the location (pose) of the goal state of an assembly operation (see Figure~\ref{fig:assembly_steps}). Furthermore, human demonstrations are provided to learn motor skills for the robot to perform assembly, which can generalize to the varying goal locations that the vision system can estimate. We use dynamic movement primitives (DMPs) to learn full $6$ DoF trajectories for learning motion profiles for the robot from demonstrations \cite{Ijspeert2002MovementRobots}. The human expert provides the task-related constraints to the robot so that the robot learns successful motor skills without the need for exploration to find suitable motor skills for the desired tasks.

The proposed collaborative assembly system is demonstrated on a robotic system which is shown in Figure~\ref{fig:assembly_steps}. The proposed system is tested for different assembly scenarios where the location of the assembly objects could be changed by a human during actual assembly. Effectively, this results in synchronization between the human and the robot via the position of the parts placed by the human. In particular, the proposed work has the following contributions:
\begin{itemize}
    \item We present a learning-based human-robot collaborative system for performing assembly.
    \item We present the design and usage of a force/torque (F/T) sensor-based kinesthetic teaching controller that simplifies interaction with the 6 DoF manipulator arm for providing demonstrations.
    \item We test the proposed system for generalization on different goal and initial locations.
\end{itemize}
As we have shown in Figure~\ref{fig:assembly_steps}, the proposed kinesthetic controller is used to provide demonstration for the insertion task which is performed by the robot. This allows the robot to perform single-shot learning for generating motion trajectories which can be parameterized by the initial and goal pose. During execution, a new motion trajectory is generated using the learned DMP and the target pose estimate provided by the proposed vision module. The robot then performs insertion while the human finishes off the assembly. 



\section{Related Work}\label{sec:related_work}
Robotic assembly is a very successful application of control and automation technology, and many manufacturing industries rely heavily on it for various products, especially in large-scale mass production. Such traditional robotic technology relies on exact placement of all parts by means of jigs and fixtures to eliminate all uncertainty so that precisely predetermined paths of the robots can be executed over and over again without any modification. Even in the absence of uncertainty, the system integration effort typically far exceeds the cost of the actual robotic hardware. In this paper, we present a collaborative assembly system which can be designed using learning from demonstration (LfD) using a vision module to detect and locate parts for assembly. Previous efforts to create collaborative human-robot systems for assembly tasks could be found in~\cite{kyrarini2019robot,wang2021optimised, 9223531}.

To allow modifications to the path of a robot in response to sensor inputs, technologies for interpretation of such inputs have been researched over many years. For assembly applications, where the assembled parts come in close contact with one another, it has been particularly necessary to react properly to contact forces, typically mounted by means of a force/torque (F/T) sensor mounted on the wrist of the robot. Various compliant control methods have been proposed and are often provided as standard control modes on industrial robot arms \cite{Whitney1987HistoricalControl}. Although methods such as impedance and admittance control can be very effective in dealing with minor misalignment, especially if the parts are chamfered, they typically cannot deal with large misalignment, e.g. of magnitude comparable with the dimensions of the assembled parts. Recent work on learning control has sought to leverage advances in deep reinforcement learning to learn adaptive non-linear control laws that can deal with such large misalignments \cite{Inoue2017DeepTasks,9838102, https://doi.org/10.48550/arxiv.2007.11646}. 

Nevertheless, if the position of assembled parts is allowed to vary in much wider ranges, for example the surface of a workbench, which human workers would often use in its entirety, force sensing alone would not be sufficient for successful assembly. In such cases, the position of the assembled parts can be identified by means of various computer vision methods. Although classical computer vision methods could use geometric models of the parts, (such as CAD files), to determine their position in an image \cite{Horn1986RobotVision}, such estimation can be brittle. A recently proposed very effective alternative is to use machine learning technology to learn the appearance of objects by generating synthetic data, and estimate the full 3D pose from this appearance by means of a deep neural network \cite{Tremblay2018DeepObjects}. Advances in depth camera imaging have further increased the accuracy of registration of point clouds to geometric shapes \cite{Li2015AModel}. However, all of these approaches require geometric information. Yet another class of methods rely on registration techniques between pairs of images (or point clouds), to recover the relative transform in 2D (respectively 3D) \cite{Yang2020Teaser:Registration}. Because such algorithms do not rely on geometric models, they might be applicable to a wider range of assembly problems.

Another technology relevant to collaborative assembly is learning from demonstration (LfD), in particular by means of novel teaching methods such as kinesthetic teaching that were too dangerous to use with traditional industrial robot arms. LfD technology (\cite{Ravichandar2020RecentDemonstration}) has been researched actively as a very appealing alternative to traditional laborious robot programming, as well as planning technology whose deployment can also be laborious and expensive, if the entire geometry of the scene has to be recreated in digital form. In particular, Dynamic Movement Primitives (DMPs, \cite{Ijspeert2002MovementRobots}) have been widely used as a parametric form of a path demonstrated by a user that can be adapted to a new starting and goal position, while retaining the general shape of the path. The demonstration can be provided by means of any method for teleoperation of the robot, and for the case of collaborative robots, hand holding, or kinesthetic teaching, could be a very natural and easy-to-use option. One popular way to implement a kinesthetic teaching controller (KTC) has been to make the robot behave like a virtual tool \cite{Kosuge1993ControlOperator}. Usability studies have shown that human operators indeed find such a virtual-tool simulating KTC easy and natural to use \cite{LopezInfante2011UsabilityInteraction}.

\section{Preliminary Material}\label{sec:background}
In this section, we present some background material for the completeness of the proposed work. For brevity, we omit many related concepts. For more details of these concepts, interested readers are refered to~\cite{schaal2006dynamic, ude2010task, https://doi.org/10.48550/arxiv.2209.14461}.
\subsection{Dynamic Movement Primitives (DMPs)}\label{subsec:DMP}
DMPs were first introduced by Schaal et al.~\cite{schaal2006dynamic}. To remove explicit time dependency, they use a canonical system to keep track of the progress through the learned behavior:
\begin{equation}
\tau \dot s = -\alpha_s s
\end{equation}

where $s = 1$ at the start of DMP execution (and $\alpha_s > 0$) and $\tau > 0$ specifies the rate of progress through the DMP.

To capture attraction behavior, DMPs use a spring-damper system (the transformation system) with an added nonlinear forcing term. Writing the DMP equations as a system of coupled first-order ODEs yields:

\begin{align}
    \tau \dot z &= \alpha_z(\beta_z (g - y) - z) + f(s)\\ 
    \tau \dot y &= z 
\end{align}

where $g$ denotes the goal pose. The forcing term is defined as a radial-basis function:

\begin{align}
    f(s) &= \frac{\sum_{i=1}^N w_i \psi_i(s)}{\sum_{i=1}^N \psi_i(s)}\\
    \psi_i(s) &= \exp{(-h_i (s-c_i)^2)}
\end{align}

where $h_i$  and $c_i$ denote the width and center of the Gaussian basis functions, respectively. The forcing term is learned from the demonstration by solving a locally weighted regression to fit the demonstration data given by the expert. Note that the formulation above describes the regular cartesian DMPs. However, we use the full $6$ DoF DMPs for generating motion of the manipulator arm during our experiments where the angular movement is modeled using quaternion formulation for the DMPs~\cite{ude2014orientation}.

\section{Proposed System}\label{sec:proposed_system}
In this section, we present the collaborative human-robot system for assembly. We explain the design and implementation of different components. 
\subsection{Design of Admittance Kinesthetic Controller (KTC)}\label{sec:kinesthetic_controller}
The proposed system makes use of learning from demonstration (LfD) for designing motion trajectories for the insertion task performed by the robot. In order to be able to efficiently design motion trajectories for the manipulator using LfD, we need to be able to provide demonstrations efficiently. This means that it should be intuitive as well as easy to move the robot while providing demonstrations. While using a joystick may provide us an easy-to-use interface for industrial robots, it becomes non-intuitive to control the full $6$ DoF pose of the end-effector for a task. Kinesthetic teaching, where a user can directly move the end-effector while applying suitable forces is a good alternative. However, designing a kinesthetic teaching controller (KTC) could be challenging for bulky, industrial robots. Another challenge is that most of the industrial robots come with position-controlled movement and they do not have torque sensors at the joints. Consequently, we present the design of an admittance-based KTC for moving the robot which makes use of a F/T sensor at the wrist of the robot. 
\begin{figure*}
    \centering
    \includegraphics[width=0.90\textwidth]{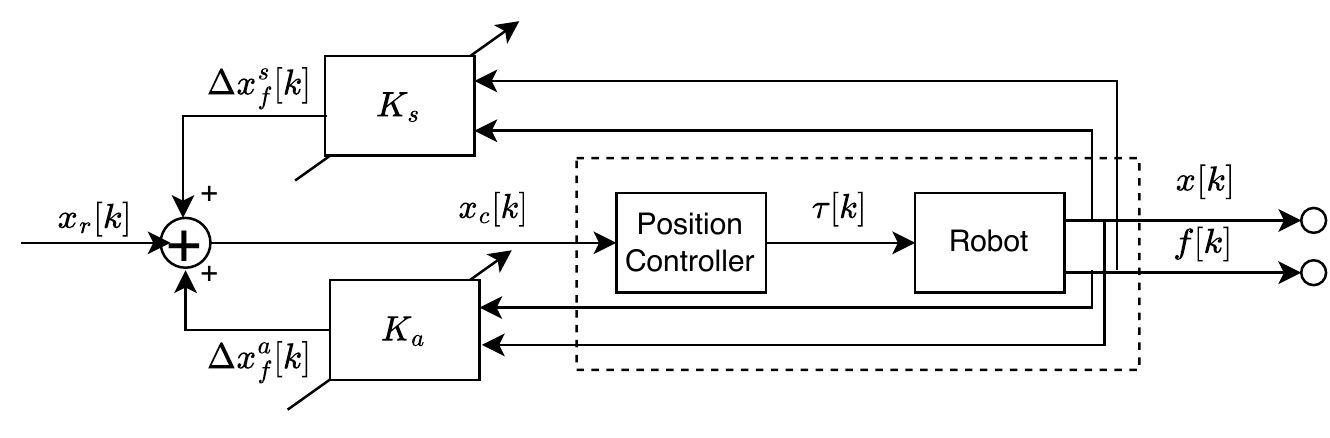} %
    \caption{The general-purpose flow diagram for the design of the kinesthetic teaching controller using force readings from the Force-Torque sensor mounted at the interface between the wrist of the robot and the actual end effector. $K_s$ denotes the stiffness matrix for the stiffness control of the robot and $K_a$ denotes the accommodation matrix.}
    \label{fig:ktc_design}
\end{figure*}

To achieve this, we use the F/T sensor attached at the wrist of the manipulator (see Figure~\ref{fig:assembly_steps}). The robot movement can be controlled based on the forces experienced by the robot which can be measured by the F/T sensor. 
We design a number of different such kinesthetic controllers with different degrees of complexities. A general-purpose flow diagram of the design of the kinesthetic controller for the robot is shown in Figure~\ref{fig:ktc_design}. As shown in Figure~\ref{fig:ktc_design}, the performance and sensitivity of the controller can be fine-tuned by tuning the stiffness matrix ($K_s$) and accommodation matrix ($K_a$) as shown in Figure~\ref{fig:ktc_design}. The proposed KTC controller can be used for providing demonstration in  both contact-free environment as well as contact-rich tasks. Furthermore, it also allows us to select the degrees of freedom which we would like to move during any demonstration. 

Since the robot we use is a position controlled robot, the KTC computes a new commanded position for the robot based on the wrench measured using the F/T sensor at the wrist of the robot. The commanded position is computed using the following law:
\begin{equation}
    x_c[k+1]=x_r[k]+\Delta x_f^s[k]+\Delta x_f^a[k], 
\end{equation}
where $\Delta x_f^s[k]=K_s^{-1}f[k]$ and $\Delta x_f^a[k]=K_a f[k]$. The stiffness and accommodation matrices ($K_s$ and $K_a$ respectively) are tuned in order to get the desired behavior for the robot. In this paper, we use diagonal stiffness and accommodation matrices (which makes design of the controller simple).


\subsection{Hole Localization Using Vision}\label{sec:pose_estimation}
We propose to use hole-detection-based insertion using vision for our approach, based on our previous work~\cite{jain2022vision}. Using traditional computer vision approaches~\cite{briechle2001template, yuen1990comparative} to detect hole location is not robust with an unknown pose of the part. We selected a supervised learning approach to detect all accessible hole locations on the part from visual sensory data obtained from an RGB-D sensor (Intel Realsense D435) camera. We implement the Mask-RCNN~\cite{he2017mask} deep learning architecture to learn masks for hole locations with instance-level segmentation. 

\begin{figure*}[ht]
    \centering
    \includegraphics[width=1.0\textwidth]{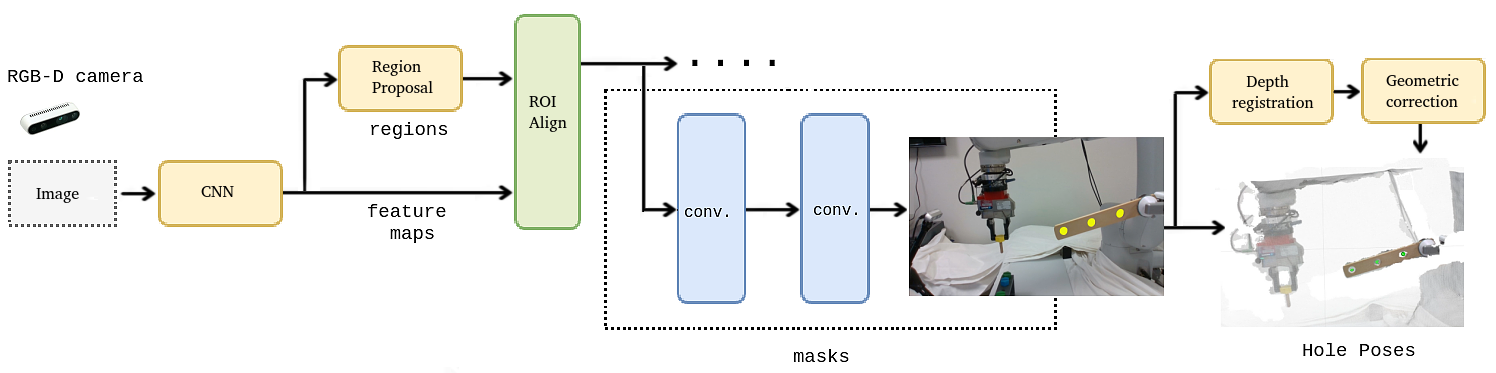} %
    \caption{The deep learning pipeline for hole pose localization. Our vision architecture is based on the Mask-RCNN model to learn masks for hole locations. Note that ROI stands for region of interest.}
    \label{fig:pipeline}
\end{figure*}

Figure~\ref{fig:pipeline} shows the vision pipeline for hole localization with the network architecture. The network architecture is built on a backbone convolutional neural network architecture for feature extraction. The backbone network could be any convolutional neural network for image analysis. We use a feature pyramid network (FPN) based on ResNet-50, which takes advantage of convolutional neural networks' inherent hierarchical and multi-scale nature to derive useful features at many different scales. Mask-RCNN lies on region proposals which are generated via a region proposal network. It follows the Faster-RCNN model of a feature extractor followed by this region proposal network, followed by an operation known as ROI-Pooling to produce standard-sized outputs suitable for input to a classifier. However, Mask-RCNN replaces the somewhat imprecise ROI-Pooling process used in Faster-RCNN with ROI-Align, which allows for accurate instance segmentation masks. It also adds a network head, a small fully convolutional neural network, to produce the desired instance segmentations. Finally, mask and class predictions are decoupled; the mask network head predicts the mask independently from the network head predicting the class. Typically, this entails using a multitask loss function $L = L_{cls} + L_{bbox} + L_{mask}$. In our scenario, we post-process the mask network head output that independently predicts the mask for learning hole localization in images. Finally, we perform geometric correction with curve fitting that has the best fit to the depth points obtained from the mask by trying to find the best visual fit of a circle.

\begin{figure}[h]
    \centering
    \includegraphics[width=0.5\textwidth]{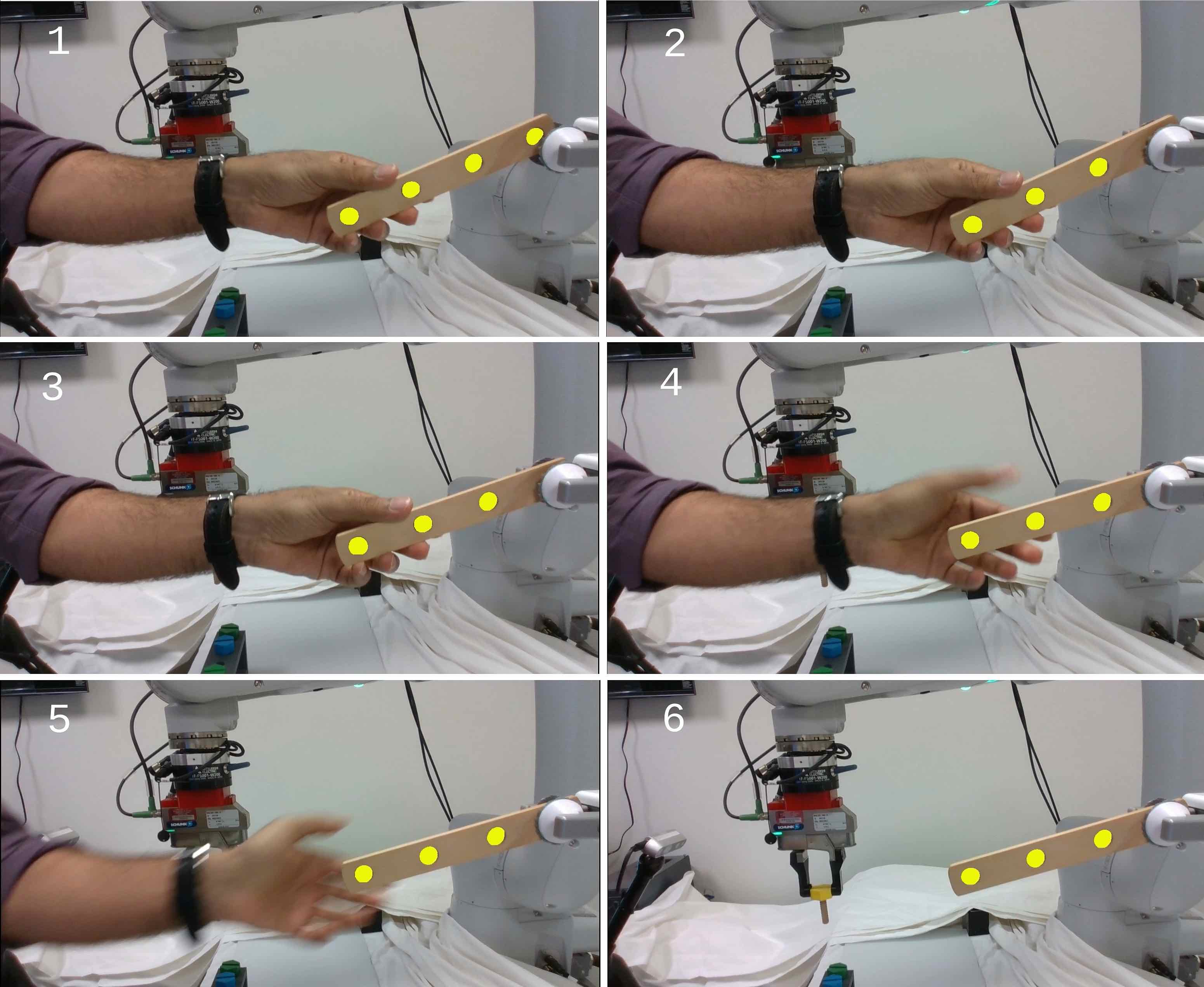} %
    \caption{Sampling of vision pipeline qualitative results for hole pose localization during part handover operation by the human operator. As seen in these pictures, the proposed hole localization method is able to detect and track the holes as the assembly part is moved in the view of the camera.}
    \label{fig:visionresults}
\end{figure}

For training the network, we performed transfer learning from the MS COCO dataset~\cite{lin2014microsoft} pre-trained weights in a supervised manner. We captured hundred images of size 640×480 at different poses, and annotated the data to indicate hole pixels with the labelme~\cite{russell2008labelme} annotation tool to train the network. We identify the resulting segmentation masks for hole locations with the network prediction. The detected segmentation mask of the hole locations are utilized with depth estimation to compute the corresponding registered point cloud data points. The output from the approach is the estimate of the hole locations from the visual sensory data for performing the insertion. Figure~\ref{fig:visionresults} shows some qualitative results of hole detection on the objects used in this particular study. This method is used to locate the goal location when the bar is grasped by the stationary gripper during the collaborative assembly.

\subsection{Trajectory Generation Using Dynamic Movement Primitives (DMPs)} One of the key requirements of a collaborative assembly system is that the robot action should be able to adapt to uncertainty in human actions, since human actions are usually far from repeatable. Thus, the robot should be able to adjust its movement depending on the result of human actions. Furthermore, it is desirable that the collaborative assembly system should avoid laborious motion planning to allow this adjustment. Thus, we make use of dynamic movement primitives to generate motion trajectories for the robot during the collaborative assembly process.

The kinesthetic controller that we had described in Section~\ref{sec:kinesthetic_controller} is used to provide a demonstration for the full $6$ DoF movement of the robot (see Figure~\ref{fig:assembly_steps}). The demonstration is used to generate trajectories for insertion of the threaded bolt using the location of the hole provided by our hole localization module described in Section~\ref{sec:pose_estimation}. The DMPs described in Section~\ref{sec:background} are provided with the tuple of initial pose and goal pose for the end-effector and a new motion trajectory is computed using the learned DMP from the demonstration. This computed motion trajectory is implemented to perform the insertion task by the robot for the assembly task. We ignore any unexpected contact or any change in pose of the peg that might happen during the task execution.

\section{Experiments}\label{sec:Experiments}
In this section, we present all the experiments we performed to verify our proposed kinesthetic controller, pose estimator, and the system for performing collaborative assembly. The experiments are performed to understand and answer the following questions:
\begin{enumerate}
    \item How does the proposed Kinesthetic Teaching Controller compare against the default built-in kinesthetic controller that does not use the F/T sensor?
    \item How well is the proposed collaborative system able to generalize for assembly?
\end{enumerate}
In the rest of this section, we will try to answer the above two questions.
\subsection{Task Description} 
\begin{figure}
    \centering
    \includegraphics[width=0.49\textwidth]{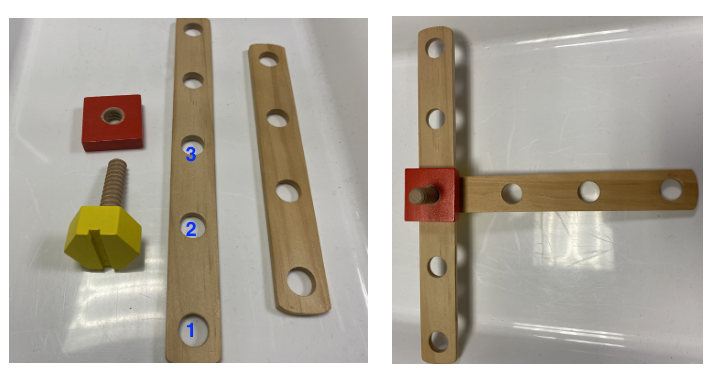} %
    \caption{The parts used  are shown in the left side figure, and the goal is to assemble the T-shaped object shown in the right side of the image. To show generalization, we use one of the three holes marked in the image at any novel position of the bar in the stationary gripper.}
    \label{fig:collab_assembly_parts}
\end{figure}

The task that we consider in this paper is that of collaborative assembly where a robot is supposed to collaborate with a human to perform a task. More specifically, the task is to assemble a desired "T"-shaped product using multiple parts. The different parts and the final desired product could be seen in Figure~\ref{fig:collab_assembly_parts}. The objective is to create the "T"-shaped object using two bars with multiple holes, a threaded bolt, and a nut. 

\begin{figure}
    \centering
    \includegraphics[width=0.49\textwidth]{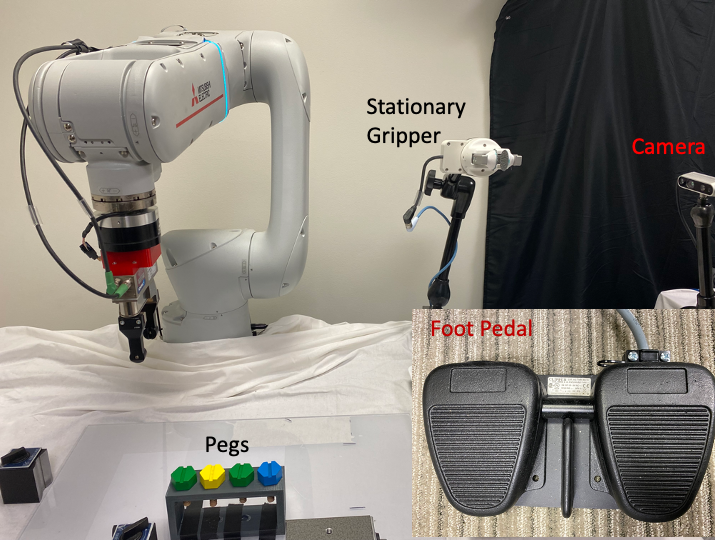} %
    \caption{In this figure, we show the assembly consisting of a stationary gripper, a collaborative $6$ DoF manipulator arm, an RGBD camera and a foot pedal (shown in the inset in the right-lower corner) which is used to indicate the end of various steps during the collaborative assembly task.  We localize the pegs with another camera (not visible in the image).}
    \label{fig:assembly_system}
\end{figure}

Since the task requires four different parts during assembly, we propose to use four dexterous hands for performing the assembly. Two of these hands are human hands, and the other two hands consist of a stationary gripper (see Figure~\ref{fig:assembly_system}) and a $6$ DoF manipulator arm mounted with a two-fingered gripper (see Figure~\ref{fig:assembly_system}). Since the full assembly task is a long-horizon, multi-step process, we need a way to indicate or infer beginning and end of the different steps of the multi-step assembly. While inferring the beginning and end of steps during the assembly is desirable, discussion of such a perception system is out of the scope of the current paper. In this work, we make use of foot pedal that is used to indicate the beginning and end of a step. All the system components are shown in Figure~\ref{fig:assembly_system}. We use a Mitsubishi Electric Factory Automation (MELFA) RV-$5$AS-D
 Assista $6$-DoF  arm (see Figure~\ref{fig:assembly_system}) for the experiments. The robot has pose repeatability of $\pm 0.03$mm. The robot is equipped with the Mitsubishi Electric F/T sensor $1$F-FS$001$-W$200$ (see Figure~\ref{fig:assembly_system}).  

In this assembly, a human first picks up a bar with holes and places it in the stationary gripper (see the initialization step in Figure~\ref{fig:assembly_steps}). Then, the task of the robot is to grasp a threaded bolt and insert it in one of the holes of the bar depending on the desired assembled product. The the human inserts the second bar into the peg (still held by the robot) and then tightens the nut on the bolt to finish the assembly (see Figure~\ref{fig:assembly_steps} to see all these steps during execution). During this multi-step assembly, a human makes use of the foot pedal to signal to the robot the end of one step and the beginning of the next one (see Figure~\ref{fig:assembly_system}).

\begin{figure}
    \centering
    \includegraphics[width=0.49\textwidth]{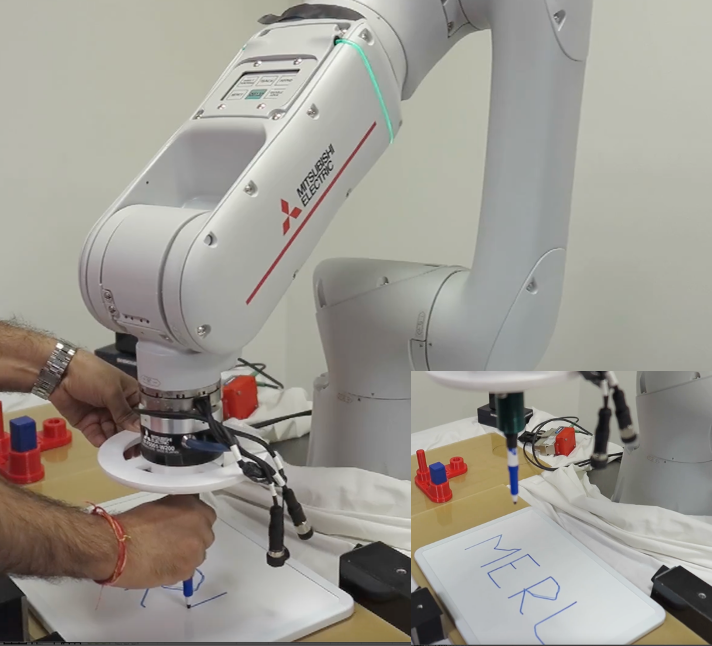} %
    \caption{The proposed kinesthetic controller can be used for demonstrating contact-rich as well as contact-free tasks. For example, in this picture we show the task of writing on a whiteboard using the proposed kinesthetic controller. The picture in the bottom-right inset shows the final word written using the KTC.}
    \label{fig:ktc_writing}
\end{figure}

\subsection{Experimental Results} In this section, we present results to evaluate the efficiency of the proposed controller. In the first set of experiments, we perform evaluation of the proposed kinesthetic controller and then perform experiments to evaluate the efficiency of the proposed collaborative system.

\subsubsection{Evaluation of the Proposed Kinesthetic Controller}\label{subsubsection:KTC_results}

In this section, we present the effectiveness of the proposed KTC for two different tasks. Since the KTC makes use of the F/T sensor, it can be used for demonstration of contact-rich tasks without running into the risk of breaking either the robot or the parts used in the task. We test the effectiveness of the KTC for a writing task. In Figure~\ref{fig:ktc_writing}, we show the use of the proposed KTC controller for a writing task. We can observe that we can get very smooth trajectories during writing (see the inset in Figure~\ref{fig:ktc_writing} for the result of writing task).  Using a suitable choice of the accommodation controller (which could be linear or non-linear and is a design choice), we can make the KTC as reactive as is required for the writing task. However, it might lead to instability as the robot may move with noise of the sensor. Thus, the KTC controller is designed as a trade-off between the sensitivity and stability for the desired task. While it is possible to find the optimal set of parameters for the accommodation as well as stiffness using data-driven optimization methods (such as Reinforcement Learning), such a discussion is out of the scope of this paper.

\begin{figure}
    \centering
    \includegraphics[width=0.54\textwidth]{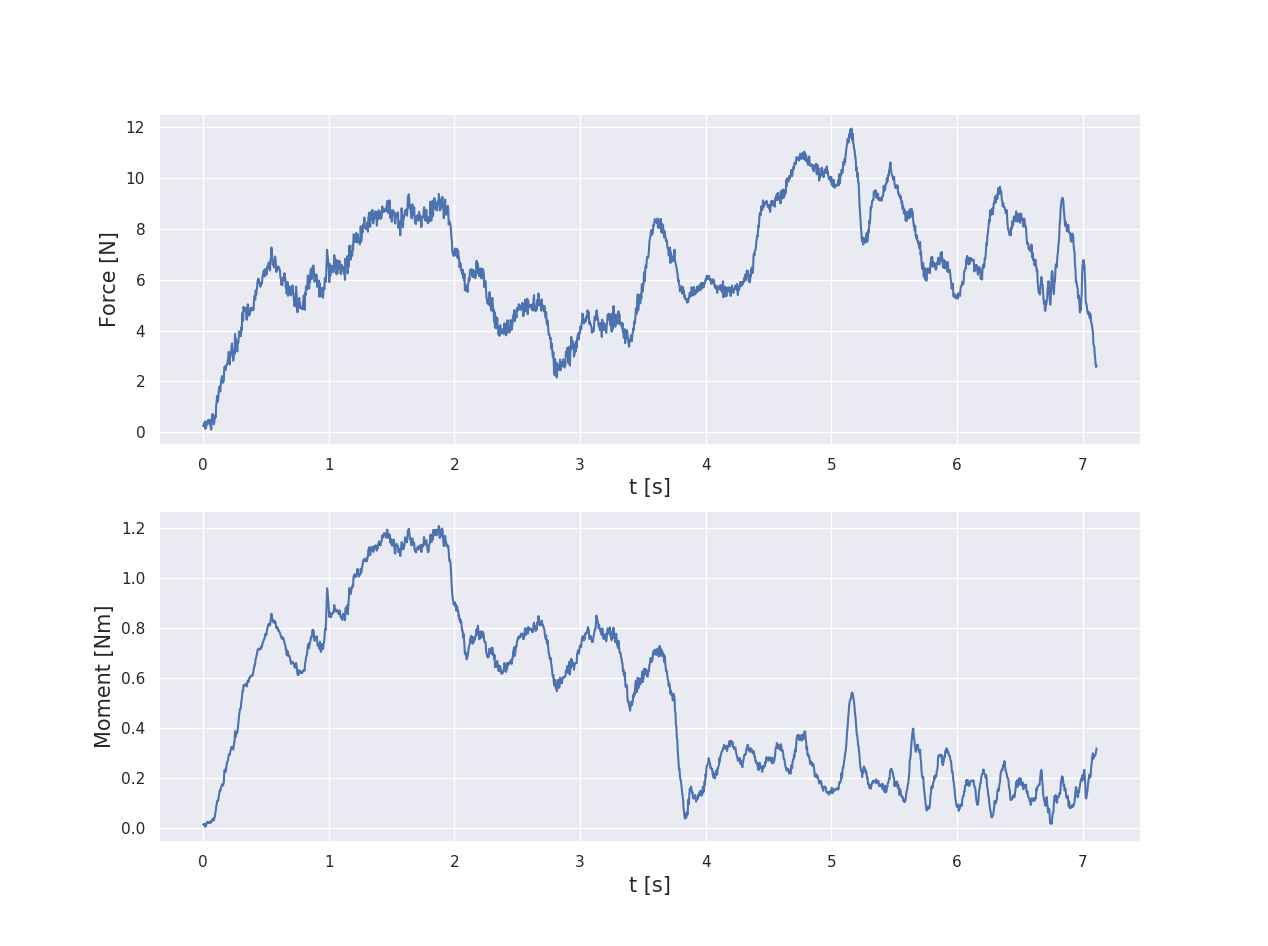} %
    \caption{The force trajectory measured by the F/T sensor at the wrist-gripper interface as a user is demonstrating a trajectory using the proposed kinesthetic controller. Despite the large moving mass of the robot, the maximum force applied by a user during the teaching motion is only $\sim$ $12$ N.  By comparison, the force required in the native cobot teach mode was in excess of 50 N (limited by instrumentation over-range) due to the need to back-drive the robot servomotors through their reduction gearboxes}
    \label{fig:ktc_force}
\end{figure}

To show the effectiveness of the proposed KTC for providing demonstrations for performing insertion, we present the force trajectory measured by the F/T sensor at the wrist of the robot. A sample force and moment trajectory for the robot during a demonstration by a user is shown in Figure~\ref{fig:ktc_force}. As could be seen in the Figure, the maximum force norm measured by the F/T sensor is about $12$ N. The force trajectory shown in Figure~\ref{fig:ktc_force} is an order of magnitude less than what an user has to apply while moving the robot using the default Kinesthetic controller that makes use of current sensors at the joints to move the robot (note that since the default kinesthetic controller does not use the F/T sensor for movement, we can not measure the exact force trajectory).

Furthermore, we collected demonstrations using the native (default) KTC and compared the timing and jerk of the demonstration trajectories against the proposed KTC. All these demonstrations were started at the same pose of the robot and were terminated after demonstrating a successful insertion task. The timing results are listed in Table~\ref{table:demo_timing} where it can be seen that the proposed KTC can smoothly move the robot and thus takes much less time compared to the default KTC (that does not make use of the F/T sensor at the wrist). To compute the jerk of the demonstration trajectories, we perform finite difference of the trajectories thrice and find the norm of jerk trajectories. The mean and standard deviation of the norm of the jerk trajectory are tabulated in Table~\ref{table:demo_trajectory_quality}. We also compute the mean and standard deviation of the maximum jerk for the demonstrated trajectories over the $5$ trajectories. As could be seen in Table~\ref{table:demo_trajectory_quality}, the trajectories demonstrated using the proposed KTC has lower jerk when compared to the default KTC for the robot.

\begin{table}[t]
    \caption{{Demonstration Timing}}
    \centering
    \begin{tabular}{c|c}
     Controller & Avg. Time over $5$ demonstrations \\
         \hline\hline  Native KTC & $24.66\pm 3.25$ [s]\\
         \hline Proposed KTC & $\mathbf{17.17\pm 0.756}$ [s]\\
    \end{tabular}
    \label{table:demo_timing}
\end{table}

\begin{table}[t]
    \caption{{Demonstration Trajectory Quality}}
    \centering
    \begin{tabular}{c|c|c}
     Controller & Jerk Norm  & Maximum Jerk During Demo \\
         \hline\hline  Native KTC & $10.55 \pm 1.11$ & $10.84 \pm 0.73$ \\
         \hline Proposed KTC &$\mathbf{6.71\pm 0.157}$ & $\mathbf{6.99 \pm 0.00}$ \\
    \end{tabular}
    \label{table:demo_trajectory_quality}
\end{table}

\subsubsection{Generalization of the Collaborative Assembly System}
\begin{figure}[h]
    \centering
    \includegraphics[width=0.49\textwidth]{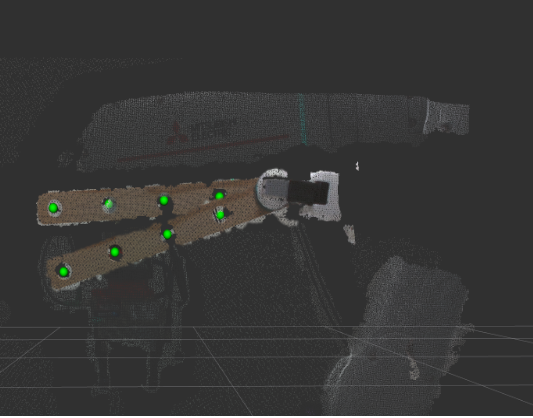} %
    \caption{In this figure, we show the range of the placement of the bar in the stationary gripper for which the vision algorithm is able to detect the location of all the four holes.}
    \label{fig:hole_detection_range}
\end{figure}

In the next set of experiments, we evaluate the generalization of the proposed collaborative system. A human user first demonstrates a single  successful insertion  of the peg into one of the three holes at a single location of the bar which is held in the stationary gripper (see Figure~\ref{fig:assembly_system} and Figure~\ref{fig:collab_assembly_parts} for location of the holes). This single demonstration is used for learning a DMP for insertion attempts during test assembly settings. 
Figure~\ref{fig:hole_detection_range} shows the angular range for the placement of the bar in the stationary gripper for which the vision algorithm is able to detect each of the three holes in the bar. To evaluate the success of the assembly, a human places the bar in the stationary gripper at the beginning of a new experiment and any of the three holes is selected for the collaborative assembly. The robot then uses the demonstrated trajectory and the position of the hole returned by the vision module to compute an insertion trajectory to perform insertion of the peg into the hole. The combination of the DMP generated motion based on the single demonstration, plus the vision algorithm's accuracy is sufficient so that the the DMP-based insertion was 100\% successful for all $20$ trials of the test run. In each of these test runs, a hole was randomly selected for insertion (out of the three possible options shown in Figure~\ref{fig:collab_assembly_parts}). It is noted that the bar was always placed in the region where the bar is in the view of the camera and thus the holes could be detected. During a few insertion attempts, the peg comes in contact with the bar during insertion, but since the robot works in stiffness control mode, it is successfully able to perform the insertion.

\section{Conclusions}\label{sec:conclusions}
Next-generation robotic systems are supposed to have an increased presence in our society where robots will have skills for perception and interacting with their environment. These robots will be used to design truly collaborative environments where robots and humans can collaborate to perform complex tasks for increased performance and efficiency. However, in order to design such a system, the robot should be able to adapt to the uncertainty in human actions. Furthermore, it is desirable that such a system could make use of human domain knowledge for performing the task and avoid laborious use of motion planning algorithms. In this paper, we presented a human-robot collaborative assembly task using DMP for motion generation, pose estimation, and force control. The proposed human-robot collaborative assembly system is designed so that the robot's movement can be adapted based on human actions using vision and LfD.

We presented an assembly task using four different parts which are put together using two human hands, one stationary gripper, and one $6$ DoF manipulator arm mounted with a two-fingered gripper and an F/T sensor. We show design and testing of an admittance-based kinesthetic controller which makes use of the F/T sensor mounted at the wrist of the robot to move the end-effector while providing demonstrations. We show the advantage of the proposed controller for moving the robot compared with the stock kinesthetic controller for the manipulator in providing demonstrations in contact-free as well as contact-rich scenarios. This kinesthetic controller is used to provide demonstrations for performing a common and useful task - the insertion of a threaded peg into a target hole during a collaborative assembly procedure. This is used to design the LfD system allowing the collaborative system to learn from human domain knowledge for the task. We showed the design and implementation of a deep learning model for hole localization which is used for performing autonomous assembly during the collaborative assembly. We show the generalization of the proposed LfD system by providing novel goal locations for insertion to the robot during testing. We show that our LfD system is able to localize the target hole locations over a reasonably large target area. 

In the future, we will work towards inferring the beginning and end of the different steps in the multi-step assembly task using vision-based techniques or using other sensing modalities. Similarly, to make the proposed assembly robust to unexpected contacts during assembly, we will make use of pose estimation using vision or tactile sensors as well as more advanced force control~\cite{9838102, https://doi.org/10.48550/arxiv.2204.10447, 9561646}, which can be used to adapt the motion trajectory to any such event during assembly. We would also like to integrate some nonprehensile manipulation capabilities so that the the system becomes modular~\cite{shirai2022robust}.

\addtolength{\textheight}{-12cm}   









\bibliographystyle{IEEEtran}
\bibliography{references,DanielsRefsFromMendeley}

\end{document}